\documentclass[sn-apacite]{sn-jnl}

\usepackage{graphicx}%
\usepackage{multirow}%
\usepackage{amsmath,amssymb,amsfonts}%
\usepackage{amsthm}%
\usepackage{mathrsfs}%
\usepackage[title]{appendix}%
\usepackage{xcolor}%
\usepackage{textcomp}%
\usepackage{manyfoot}%
\usepackage{booktabs}%
\usepackage{algorithm}%
\usepackage{algorithmicx}%
\usepackage{algpseudocode}%
\usepackage{listings}%

\theoremstyle{thmstyleone}%
\theoremstyle{thmstyletwo}%

\theoremstyle{thmstylethree}%

\begin{document}

\title[Article Title]{A Big Data Analytics System for Predicting Suicidal Ideation in Real-Time Based on Social Media Streaming Data}


\author[1,2]{\fnm{Mohamed A.} \sur{Allayla}}\email{mohamed.abdulstar@uomosul.edu.iq}

\author*[2,3]{\fnm{Serkan} \sur{Ayvaz}}\email{seay@mmmi.sdu.dk}


\affil[1]{\orgdiv{Dams and Water Resources Research Center}, \orgname{University of Mosul}, \city{Mosul},\country{Iraq}}

\affil[2]{\orgdiv{Department of Computer Engineering},\orgname{Yildiz Technical University}, \city{Istanbul},\country{Turkey}}
\affil[3]{\orgdiv{Centre for Industrial Software}, \orgname{University of Southern Denmark}, \city{Sønderborg},\country{Denmark}}





\abstract{Online social media platforms have recently become integral to our society and daily routines. Every day, users worldwide spend a couple of hours on such platforms, expressing their sentiments and emotional state and contacting each other. Analyzing such huge amounts of data from these platforms can provide a clear insight into public sentiments and help detect their mental status. The early identification of these health condition risks may assist in preventing or reducing the number of suicide ideation and potentially saving people’s lives. The traditional techniques have become ineffective in processing such streams and large-scale datasets. Therefore, the paper proposed a new methodology based on a big data architecture to predict suicidal ideation from social media content. The proposed approach provides a practical analysis of social media data in two phases: batch processing and real-time streaming prediction. The batch dataset was collected from the Reddit forum and used for model building and training, while streaming big data was extracted using Twitter streaming API and used for real-time prediction. After the raw data was preprocessed, the extracted features were fed to multiple Apache Spark ML classifiers: NB, LR, LinearSVC, DT, RF, and MLP. We conducted various experiments using various feature-extraction techniques with different testing scenarios. The experimental results of the batch processing phase showed that the features extracted of (Unigram + Bigram) + CV-IDF with MLP classifier provided high performance for classifying suicidal ideation, with an accuracy of 93.47\%, and then applied for real-time streaming prediction phase.}




\maketitle

\section{Introduction}\label{sec1}
Suicidal ideation is a serious public health concern. The number of suicidal ideations is increasing at an alarming rate every year. According to a report issued by the World Health Organization (WHO), more than 703,000 people commit suicide annually, which means roughly one person dies every 45 seconds due to suicide. In addition, 25 suicide attempts for each suicide, and many more had serious thoughts about suicide \cite{rf1}. Suicidal ideation has continuously been linked to emotional states such as depression and hopelessness \cite{rf2}. The early detection of suicidal ideation may help to prevent many suicide attempts and identify individuals needing psychosocial support.  

Traditional methods and programs for suicide prevention are still reactive and require patients to take the initiative to seek medical help. However, many patients are not highly motivated to receive the necessary support. Owing to the anonymity characteristic of online social media platforms, it has become an alternate space for people to express their honest feelings or thoughts about their suffering or healthcare issues without fearing stigma or revealing their real identities, as in face-to-face conversation \cite{rf45}.

This is considered a valuable source for detecting high-risk suicidal ideations instances and uncovering these dangerous intentions before they become irreversible or the sufferers end their lives. Suicidal sufferers may show suicidal intentions online through brief ideas or detailed planning. Social media have been successfully leveraged to assist in detecting physical and mental illnesses more easily \cite{rf4}. Therefore, researchers have begun using online postings to detect suicidal ideation manually or with the help of machine learning techniques \cite{rf5}. Manual identification of suicidal ideation has become more challenging due to the vast amount of content on social media platforms. 

Moreover, social media posts are generated as streaming data in real-time. However, real-time systems require direct input and rapid processing capability to make decisions in a short time \cite{rf6}. Several problems must be addressed before developing a real-time analytics system. The first is to provide a reliable and efficient framework for distributing data without losing accuracy. Most big-data research in healthcare focuses on the technical aspects of big data. Apache Spark \footnote{https://spark.apache.org/} and Apache Kafka \footnote{https://kafka.apache.org/} are examples of these frameworks. Another problem with streaming data is that it involves high-velocity and continuous data generation. Hence, processing such a huge data stream using a traditional system environment in real-time may result in system bottlenecks. 

The presented work aimed to build an effective real-time model using a big data analytics system to predict a person’s suicidal ideation at an earlier stage based on their social media posts. We focused primarily on a social media platform where people talk about different mental health issues and offer a platform to help. Reddit’s topic-specific subreddits were used as the source of the historical data utilized to train batch processing phase classifiers. The Reddit data contains postings from subreddits titled “Suicide Watch” and “Teenagers.” In addition, we extracted streaming tweets using Twitter API for real-time prediction. The API offers a collection of ways to communicate with the application. We employed spark-structure streaming to handle the Streaming of tweets in real time. 

Some notable contributions made by this paper include the following:
\begin{itemize}
    \item This paper proposed a scalable predictive system that can analyze large volumes and high-velocity streaming data in real-time using “big data” architecture to predict suicidal ideation cases that require special attention.
    \item We applied various experiments with multiple Apache Spark ML algorithms using three feature extraction: TF-IDF, N-gram, and CountVectorizer, with various combinations and testing scenarios.
    \item We performed optimization techniques to achieve high prediction accuracy. The proposed system achieved significant performance on both batch and real-time streaming phases of suicidal ideation prediction. 
\end{itemize}

The remainder of the paper is structured as follows: Section \ref{sec2} briefly reviews the literature on detecting suicidal ideation. Section \ref{sec3} describes the proposed methodology and big data architecture that is used for detecting suicidal ideation. Section \ref{sec4} presents the experimental setup, including a performance analysis comparison of the supervised classification algorithms used with various testing-type scenarios and an exploration of the data analysis. Section \ref{sec5} discusses the results obtained and the notable findings of the work. Finally, Section \ref{sec6} concludes the paper and describes the future work.

\section{Literature Review}\label{sec2}
Sentiment Analysis has attracted the attention as a research topic in various fields such as financial \cite{rf41}, public health \cite{rf44}, product reviews \cite{rf42}, voting behavior \cite{rf43}, political \cite{rf40} and social events \cite{rf16}. Although approaches, methods and models vary across domains, it has been observed that sentiment analysis and prediction tasks often produce useful and interesting results. From the perspective of monitoring suicidal ideation and mental state, there are some studies that analyze social media data using natural language processing (NLP) and sentiment analysis by investigating different aspects \cite{rf9,rf11,rf15}.

In the study conducted by S. Jain et al.\cite{rf9}, two datasets were used to develop a machine learning-based method for predicting suicidal behaviors depending on the depression stage. The first dataset was collected by creating a questionnaire from students and parents and then classifying the depression according to five severity-based stages. The XGBoost classifier reported a maximum accuracy of 83.87\% in this dataset. The second dataset has been extracted from Twitter. Tweets were classified according to whether the user had depression. They found that the Logistic Regression algorithm exhibited the highest performance and achieved an accuracy of 86.45\%.

N. Wang et al. \cite{rf10} proposed a deep-learning (DL) architecture as well as evaluated three more machine learning (ML) models to analyze the individual content for automatically identifying whether a person will commit suicide within 30 days to 6 months before the attempt. They created and extracted three handcrafted feature sets to detect suicide risk using the three-phase theory of suicide and earlier work on emotions and pronouns among people who exhibit suicidal thoughts.

R. Sawhney et al. \cite{rf11} proposed a new supervised method for identifying suicidal thoughts using a manually annotated Twitter dataset. They used a set of features to train the linear and ensemble classification algorithms. The most significant contribution of their work was the performance enhancement of the Random Forest algorithm compared with other classification algorithms and baselines. Comparisons also were made with baseline models applying different methodologies, including LSTMs, negation resolution, and rule-based approaches. Their work proved that the Random Forest algorithm outperformed the other classifiers and baselines.

Similarly, M. Chatterjee et al. \cite{rf46} analyzed Twitter platform content and identified the features that can hold signs of suicidal ideation. Multiple ML algorithms were applied, including LR, RF, SVM, and XGBoost, to evaluate the effectiveness of the suggested approach. The study involved extracting and combining various topics, linguistic, statistical features, and temporal sentiments. The study extracted multiple features from Twitter data, including sentiment analysis, emoticons, statistics, TF-IDF, N-gram, temporal features, and topic-based features (LDA). The empirical findings showed that by employing the Logistic Regression classifier, an accuracy of 87\% was registered.

A. E. Aladağ et al. \cite{rf13} used text mining implemented on post titles and bodies; they built a classification model that differentiated between postings that were suicidal and others that were not suicidal. The needed features were extracted using various techniques, including TF-IDF, word count, linguistic inquiry, and sentiment analysis of the titles and bodies of the posts. In addition, several classification algorithms were applied. The suicidality of posts was correctly classified using Logistic Regression (LR) and Support Vector Machine (SVM) classifiers. Accuracy and an F1 score were achieved at 80\% and 92\%, respectively.

By using data collected from electronic medical records in mental hospitals, N. J. Carson et al. \cite{rf14} built and evaluated an NLP-based machine-learning approach to detect suicidal behaviors and thoughts among young people. 

A. Roy et al. \cite{rf17} evaluated psychological weight factors, including depression, hopelessness, loneliness, stress, anxiety, burdensomeness, and insomnia. Furthermore, the sentiment polarity and Random Forest (RF) algorithm were applied with ten estimated psychological measures for predicting SI within tweets and achieved an 88\% AUC score.

On the other hand, V. Desu et al. \cite{rf15} proposed an approach that utilizes various ML and DL algorithms, such as XGBoost, SVM, and ANN, implemented upon a Spark cluster with multiple nodes to detect individuals who suffer from depression and suicidal thoughts and require urgent assistance or support by analyzing their social media content. The proposed ANN model provided superior efficacy over all other baseline algorithms and registered the best accuracy rate of 76.80

M. J. Vioules et al.\cite{rf18} developed a novel method that uses Twitter data to identify suicide warning signs in users and detect postings containing suicidal behaviors. The key contribution of their method is its ability to detect sudden changes in users’ online behavior. To identify these changes, they employed NLP algorithms with a martingale framework to collect behavioral and textual features. The experimental results demonstrated that their text-scoring method could detect warning signs in a text more effectively than standard machine learning classifiers.


W. Jung et al. \cite{rf19} designed multiple machine-learning models and analyzed suicidality using Twitter data. The models were trained using 1097 suicidal and 1097 nonsuicidal tweets. They explored metadata and text-feature extraction to construct efficient prediction models. They trained the classifier models using Random Forest and Gradient-boosted tree (GBT). The experiments were conducted using multiple features to construct a robust classifier. The model achieved good accuracy, with an F1-Score of 84.6\%.

M. M. Tadesse et al. \cite{rf20} used NLP techniques to identify the depressive content of users generated on the Reddit social website. The study primarily focused on deploying and evaluating several feature extraction approaches, such as LIWC, N-grams, and topic modeling utilizing LDA to achieve the highest performance results. The authors applied several classification algorithms, including LR, SVM, RF, Adaptive Boosting (AB), and Multilayer Perceptron (MLP), to evaluate the risk of depression among users. The experimental results were maintained in a confusion matrix to measure the model’s performance. The Multilayer Perceptron (MLP) model showed high effectiveness with LIWC, Bi-gram, and LDA features combination, which resulted in the most outstanding performance for depression identification at 91\% accuracy with an F1 score of 93\%.

M. Birjali et al. \cite{rf21} provided a method for building a suicide vocabulary to solve the lack of current lexical resources. To improve their analysis, they proposed investigating the use of Weka as a data mining tool by using machine learning methods to gain valuable insights from Twitter data gathered via Twitter4J. The dataset was built using 892 tweets. They also introduced an algorithm that utilizes WordNet to perform semantic analysis regarding the train set and the tweets’ data set, allowing practical semantic similarity computation. They have demonstrated the efficacy of using a machine-learning-based technique through Twitter content as a suicide prevention method. The empirical findings showed that Naïve Bayes algorithms achieved a Precision value of 87.50\%, a Recall value of 78.8\%, and F1. value of 82.9\%.

N. A. Baghdadi et al. \cite{rf5} presented a detailed framework for text content classification, specifically for Twitter content. The trained model was employed to identify the tweets as “Suicide” or “Normal.” The dataset contains 14,576 tweets. Additionally, the study provided preprocessing methods specifically designed for Arabic tweets. The dataset was annotated through multiple annotators, and the framework’s effectiveness was evaluated using various assessment methods. Valuable understandings were gained through the Weighted Scoring Model (WSM). Both USE and BERT classifier models were also explored. The WSM models registered the highest-weighted sum of 80.20\%.

\section{Proposed Methodology}\label{sec3}
Real-time streaming analysis of social media content can provide helpful and up-to-date information on individuals with mental health problems. The current analytics methods that analyze social media content with massive volume offline are not robust and active for supporting real-time decision-making under essential conditions. Thus, these analysis methods must built to provide effective stream real-time prediction. The methodology comprises two phases: batch processing and real-time streaming prediction. 

\begin{figure}[h]
\centering
\includegraphics[width=0.9\textwidth]{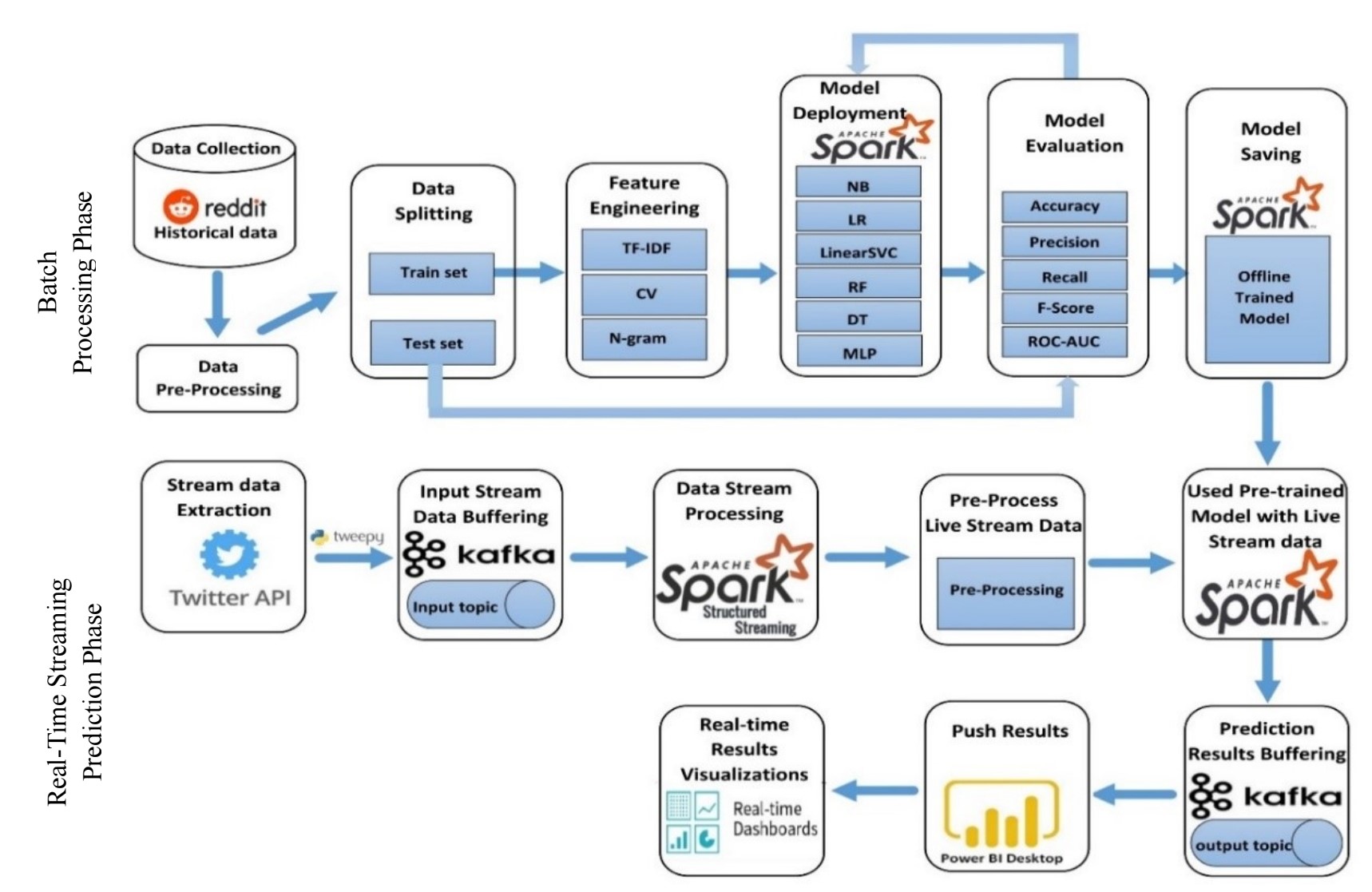}
\caption{Proposed methodology for predicting suicidal ideation on social media content}\label{fig1}
\end{figure}

Our system methodology was built based on four primary components: the input source system, where the system obtains the stream data (Apache Kafka); the stream data processing, where the stream data are processed (Apache Spark Structured Streaming); building the classification algorithms (Apache Spark ML); and the sink node, where the final results are analyzed and visualized (Power BI). We built several Apache Spark ML models using multiple feature extraction techniques. Also, we compared the classification performance of multiple models using various evaluation methods to determine the optimal architecture for predicting suicidal-related posts from real-time Twitter streaming data.
Figure \ref{fig1} provides a clear overview of the proposed methodology and the experimental workflow used in this work.

\subsection{Big Data Architecture}\label{}

This section describes the big data architecture applied in this work. Our proposed methodology was developed to efficiently analyze massive volumes of social media content with high velocity in real-time streaming data using a distributed big data environment. 

\subsubsection{Apache Spark}\label{subsec2}
Apache Spark has been applied in the proposed methodology as a data processing engine. It is an analytics platform that supports batch and stream data processing \cite{rf22}. Spark is a cluster computing system designed to be open source with various scalable and distributed ML built-in libraries \cite{rf23}. A key feature of Spark is its scalability, which enables building spark clusters with several nodes. It employs a master-slave design consisting of a Driver program that operates as the cluster’s master node and a set of executors that act as worker nodes. The core components of Spark include Spark SQL, which is used for structured query language (SQL), and Spark Streaming, which is used to process stream data. Spark Structured Streaming is developed on top of Spark SQL. Structured Streaming manages its execution incrementally and continuously, changing the final output whenever new data streams are received. 

\subsubsection{Apache Kafka}\label{}
Apache Kafka has been used to develop real-time prediction pipelines and stream data messaging. Kafka is an open-source and widely powerful ingestion system primarily used in big data applications \cite{rf24}. It is a low-latency, high-throughput system for managing and transferring massive and high-velocity data in a streaming manner. Producer and consumer API are the two primary components of the Kafka architecture. The Producer API allows the system to send data to the Kafka topics. The Consumer API provides access to Kafka topics and processes the data streams in real-time at any time. 

\subsection{Batch Data Processing Phase}\label{}
The experiments performed during the batch processing phase aimed to develop and train multiple Spark ML models with different feature extraction and testing scenarios. The model with the highest performance was then applied for real-time streaming data prediction phase. The batch processing phase consists of seven primary stages: (i) Data Collection, (ii) Data preprocessing, (iii) Dataset Splitting, (iv) Feature Engineering, (v) Model Development, (vi) Models Evaluation, and (vii) Model Saving. The upcoming subsections will provide a detailed description of each phase’s steps.

\subsubsection{Datasets Collection}\label{}

Datasets play an essential role in any text-data analysis. The dataset required for our experiment in the batch processing phase was gathered and acquired from Reddit social media platforms. The primary source of batch datasets is the Kaggle website, a publicly accessible benchmark dataset for various applications \cite{rf25}. The obtained dataset was utilized to train and assess the classifier models during the batch processing phase. The dataset was organized in a separate CSV file format and contained posts from Reddit’s platform from subreddits titled “Suicide Watch” and “Teenagers Forum,” which were collected using the’ Pushshift’ API. The dataset comprised approximately 232,074 posts collected between Dec. 16, 2008, and Jan. 2, 2021, of 116,037 were classified as suicidal and 116,037 as nonsuicidal. We cleaned and preprocessed the dataset to remove duplicate posts, empty rows, and unnecessary columns. After the preprocessing step, the dataset resulted in 232,042 rows, including 116,028 suicidal and 116,014 nonsuicidal instances. For our task, we used only the post content and target columns for the analysis task. Some batch data samples are presented in Table \ref{tab1}.

\begin{table}[h]
\caption{Samples of the Batch Dataset Postings}
\begin{tabular}{cc}

\textbf{class type} & \textbf{\textit{postings}}\\
  \hline 
\multirow{3}{*}{\textbf{Suicide}} 
&I need help just help me im crying so hard. \\
&I have nothing to live for. My life is so bleak.\\
&Suicidal tics and intrusive anxiety...\\
\hline
     
\multirow{3}{*}{\textbf{Non-suicide}} 
&I just got a Russian Hardbass song in my Spotify...\\
&I wish I could change my name to Seymour...\\
&My life is not a joke Jokes have meaning.\\
\hline
\end{tabular}
\label{tab1}
\end{table}

\subsubsection{Data Preprocessing}\label{}
\begin{figure}[h]
\centering
\includegraphics[width=0.4\textwidth]{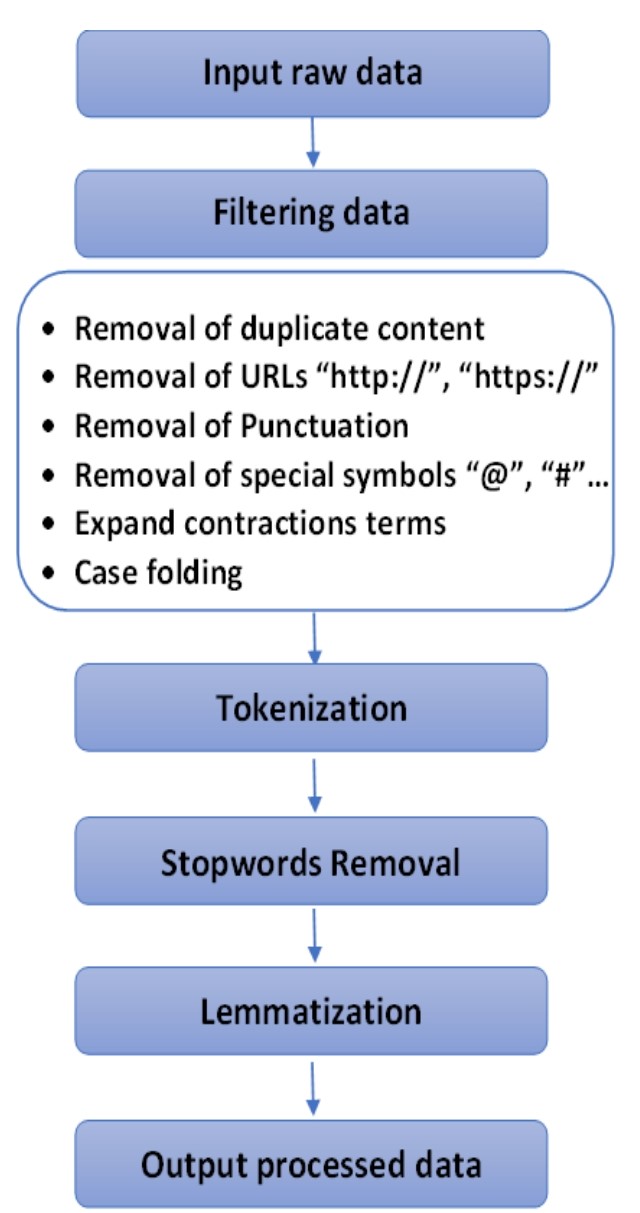}
\caption{The primary steps in preprocessing the raw dataset}\label{fig2}
\end{figure}

The text analysis performance can be improved by selecting the proper data preprocessing strategy since the input data collected from social media may contain many non-meaning words or characters, which can increase the complexity of the analysis. Hence, we aimed to prepare and refine the raw data into a suitable and understandable format for each classifier model. Some preprocessing methods are standard for text-analyzing tasks, while others depend on the complexity of data and affect the final result. We preprocessed and prepared the dataset using Natural Language Processing (NLP) techniques before passing it to the feature extraction and training stages. The preprocessing steps used to prepare the raw data were performed as illustrated in Figure \ref{fig2}.

\subsubsection{Filtering Data}\label{}

In this step, we filtered the obtained tweets to remove duplicate content, URL links (“https://," “http://,”), punctuation (e.g., “?”, “!”), special symbols (e.g., “\$”, “\%”,”\@”)  and the hashtag (“\#”). The filtering step also includes case folding and expanding contractions with their corresponding complete form (i.e., “let's" into ``let us", ``didn't" into ``did not."). This step has a significant effect on improving the effectiveness of the classifiers as it reduces the dataset complexity. 

\subsubsection{Tokenizing}\label{}
The tokenization step is essential for any natural language processing (NLP) pipeline. It has a considerable influence on the remaining phases of the pipeline. It breaks down the text data into individual, more meaningful terms, including words, punctuation marks, symbols, and abbreviations, to make data exploration more accessible. The result of this process is known as a token \cite{rf26}. These tokens were then used as input data for the processing pipeline.

\subsubsection{Stopword Removing}\label{}

This step excludes words that have no sentimental effect in the dataset. Stop words are the most frequently used term in the document. So, in this stage, we eliminated most frequently stopwords, such as pronouns like ``she" and ``he" articles such as ``and," ``the," ``a," ``an," and prepositions like ``on," ``of," ``to," ``but," ``for." and so on, therefore, in this way, we aimed to reduce the size and complexity of the dataset.

\subsubsection{Lemmatizing}\label{}
The input data was lemmatized at this step. Lemmatizing removes inflectional ends and returns each word in the dataset to its basic or dictionary form. Lemmatizing requires a comprehensive vocabulary and morphological analysis to lemmatize the words. Among various lemmatization methods, we focused on rule-based approaches using ``WordNetLemmatizer."  It employs a pre-established set of morphological and syntactic rules to find the lemma of each word within the input text. The use of Lemmatization helps to reduce the dimensionality and the vocabulary size of textual data, which leads to improved performance of analytical techniques.

\subsubsection{Dataset Splitting}\label{}
To train the classification models, it is necessary to split the dataset. Therefore, we divided the entire historical Reddit data into two subsets: Out of 80\% of the dataset applied for training data, the remaining 20\% were unseen data and applied for testing data. The classification models were trained and optimized using the training data to determine the most accurate features. On the other hand, the testing data (unseen data) was employed to assess the effectiveness of the classification models. Table \ref{tab2} provides descriptive statistics for the testing and training sets.

\begin{table}[h!]
\centering
\caption{Training and Testing Dataset Statistics}
\begin{tabular}{|c|c|c|}
\hline
\textbf{Data Subset} & \textbf{Class Type} & \textbf{No. of postings} \\ \hline
\multirow{2}{*}{Train set} & Suicide & 92726 \\ 
                            & Non-suicide & 92704 \\ \hline
\multirow{2}{*}{Test set}  & Suicide & 23302 \\ 
                            & Non-suicide & 23310 \\ \hline
\end{tabular}
\label{tab2}
\end{table}

\subsubsection{Feature Engineering}\label{}
Once we had a clean data corpus from the previous stages, it was fed into the different feature engineering methods. Our goal was to find the optimal features that provide the highest classification performance, reduce the complexity, and speed up the data transformation. In this stage, we have used three feature engineering techniques to obtain and extract the dataset's essential features, including N-gram, TF-IDF, and CountVectorizer (CV) with multiple combinations.\\

      N-gram:
      N-gram is a feature extraction method identifying N successive word groups within a text \cite{rf27}. This method is widely used as a feature extraction and analysis tool in NLP and text mining. It involves converting the input data into a series of n separate tokens. In our work, the most important features are represented using Unigrams (single words) and Bi-grams (two words have different meanings when combined) with the help of the PySpark library. Also, we assigned high importance to N-grams that appear more than four times in the document.\\

	TF-IDF:
      TF-IDF is a statistical method to extract relevant features from textual data input. TF-IDF builds a vector matrix to demonstrate a word's importance in the document. A word with fewer occurrences in a document is more appropriate for classification. TF-IDF provides a lower score for the most frequent terms and a higher score for lower-frequency terms in a document \cite{rf28} \cite{rf29}. The Spark ML API provides two methods for calculating term frequencies: HashingTF and CountVectorizer (CV). TF-IDF is calculated using the equations \ref{eq1}, \ref{eq2} and \ref{eq3} as below.

\begin{equation}
TF(t) = \frac{\text{No. of times term } t \text{ appears in a document)}}{\text{Total No. of terms in a document}}
\label{eq1}
\end{equation}

\begin{equation}
IDF(t) = \log \left( \frac{\text{Total documents No.}}{\text{No. of documents that contain the term } t} \right)
\label{eq2}
\end{equation}

\begin{equation}
TF\_IDF(t) = TF(t) \times IDF(t)
\label{eq3}
\end{equation}

CountVectorizer (CV):
It is a basic method for tokenizing data and generating a numerically-representative wordlist \cite{rf30}. It builds several columns depending on the occurrence of a unique word in the vocabulary. These columns represent each row by replacing words with their frequencies. CV can be employed when a prior dictionary is unavailable to extract the vocabulary and build the required dictionary \cite{rf31}. 
As part of this study, we conducted the experiments using the following combinations of feature extraction methods:\\ 
	Unigram + TF-IDF
	Unigram + CV-IDF
	Bigram + CV-IDF 
	(Unigram + Bigram) + CV-IDF

\subsubsection{Models Development}\label{}
In our proposed methodology, we built the classification models using multiple Spark ML algorithms, namely Naïve Bayes (NB), Logistic Regression (LR), Linear Support Vector Classifier (LinearSVC), Decision Tree (DT), Random Forest (RF), and Multilayer Perceptron (MLP) classifiers. The classifier models were trained and tested with various parameter and feature extraction combinations until the best performance values were achieved.\\

Naïve Bayes Classifier (NB):
NB is a well-known machine learning classification algorithm based on supervised learning. The NB classifier implies that the attributes are independent of each other and that the presence or absence of one attribute does not affect the other attributes. The Naïve Bayes algorithm builds based on Bayes'' theorem \cite{rf32}. The NB classifier is often used and ideal for text classification challenges due to its simplicity and speed \cite{rf33}.\\

Logistic Regression Classifier (LR):
LR algorithm is commonly employed for classifying problems and belongs to the generalized linear model category. Another term for logistic regression is the maximum entropy algorithm. LR can help calculate and predict the likelihood of allocating a new sample to a particular category for binary or multiclass classification tasks. The algorithm performs well on linearly separable datasets and can be applied to determine the correlations within dataset attributes.\\

Linear Support Vector Classifier (LinearSVC):
The LinearSVC classifier is a standard algorithm often used for large-scale classification tasks. Despite its flexibility, it is mainly used in ML to handle classification tasks. Linear SVC is a non-probabilistic classification model that needs an extensive training set. It uses a hyperplane that optimally splits the classes represented in a high-dimensional field space. LinearSVC is widely known for its practical abilities, mainly in dealing with real-world data, which include a solid theoretical basis and insensitivity to high-dimensional data.\\

Decision Tree Classifier (DT):
Decision Tree algorithm is a common machine-learning method categorized as a non-parametric supervised algorithm \cite{rf34}. It is a hierarchical model designed as a tree structure. DT is typically composed of multiple levels beginning from the root node. Every interior node holds at least one child, representing the evaluation of an input feature or variable. Based on the results of a decision test, the branching procedure will repeat itself, directing the corresponding child node along the suitable path, and this process continues until the last leaf node. The optimal tree is the shortest tree that can correctly categorize all data points and has the fewest splits.\\

Random Forest Classifier (RF): It is a popular and widely applied ML method that may be utilized or adopted for both classification and regression purposes. It was introduced by L. Breiman \cite{rf35}. RF algorithm is sometimes called a ``forest of decision trees." RF algorithm decreases the prediction variance a decision tree generates and improves its performance. For this purpose, many decision trees were merged using a bagging aggregation technique \cite{rf36}. RF learns in parallel from numerous decision trees made at random, trained on different data sets, and uses various features to get at its individual decisions. RF is more accurate and reliable than the decision tree since the final decision depends on the average of the decision tree's outputs \cite{rf37}.\\

Multilayer Perceptron Classifier (MLP): It is a form of feedforward neural network. MLP employs backpropagation, a supervised learning approach. MLP includes three sets of nodes: the first set is input-layer neurons, the second set is hidden-layer neurons, and the last set is called the output-layer neurons, which represent the final results of the system. Neurons in a perceptron require an activation function that applies a threshold, such as a sigmoid or ReLU. Any arbitrary activation function can be applied to neurons in the Multilayer Perceptron.

\subsubsection{Models Evaluation}\label{}

The performance of the proposed architecture was evaluated using various assessment methods, including Accuracy (ACC.) (Equation \ref{eq4}), Precision (PRE.) (Equation \ref{eq5}), Recall (REC.) (Equation \ref{eq6}), F1-scores (F1) (Equation \ref{eq7}), and the ROC-AUC. Furthermore, the k-fold Cross-Validation approach was employed to ensure the models fit properly without overfitting and underfitting issues. Each classifier was evaluated by calculating the average accuracy of the 10-fold cross-validation to achieve a better model performance. Table \ref{tab:3} depicts the confusion matrix form for binary classification. The confusion matrix's rows and columns are the counts of the post numbers that were either true positives (TP), False Positives (FP), True Negatives (TN), or False Negatives (FN). 
Where:\\

TP: Model classified positive class for a post, and the actual post class is also positive.\\
TN: Model classified a negative class for a post, and the actual post class is also negative.\\
FP: Model classified positive class for a post, whereas a post is negative.\\
FN: Model classified negative class for a post where a post is positive.\\

Accuracy: It is the most popular and straightforward way of measuring the model's performance. Accuracy is the ratio of samples that have been properly classified compared to the whole number of samples, as shown in Equation \ref{eq4}:

\begin{equation}\label{eq4}
Accuracy = \frac{TP + TN}{TP + FN + TN + FP} *100\%
\end{equation}\\

Precision (Specificity): It is the True-Positives (TP) ratio correctly predicted to the overall number of positively predicted samples (TP+FP). It is considered as TNR (True Negative Rate). It is calculated as in equation \ref{eq5}:

\begin{equation}\label{eq5}
Precision=Specificity = \frac{TP}{TP  + FP} *100\%
\end{equation} \\

Recall (Sensitivity): It is the ratio of positively identified observations (TP) to the overall number of positively identified observations (TP + FN). The recall is sometimes referred to as TPR (True Positive Rate). It is calculated as in equation \ref{eq6}:

\begin{equation}\label{eq6}
Recall=Sensitivity = \frac{TP}{TP  + FN} *100\%
\end{equation}\\

F1-Score: It is the average of precision and recall scores. Using F1-score assessment metrics, we can evaluate an ML classifier's performance on all data classes. F1-Score can be defined as the equation \ref{eq7}.

\begin{equation}\label{eq7}
F1-Score = 2* \frac{Precision*  Recall}{Precision+ Recall} *100\%
\end{equation}\\

ROC-AUC: It is a two-dimensional graph that employs TPR and FPR to display the ability of a classifier. In the graph, the X-axis demonstrates TPR, whereas the Y-axis demonstrates FPR. A higher AUC score indicates a more effective classification model. Ideally, the ROC curve should reach the top-left corner of the graph, resulting in an AUC score approaching 1. The AUC range score can be from 0.5 (Unsatisfactory classifier) to 1.0 (Outstanding classifier).

\begin{table}[h!]
\centering
\caption{Confusion-Matrix Structure}
\begin{tabular}{|c|c|c|}
\hline
\multirow{2}{*}{Predicted values} & \multicolumn{2}{c|}{Actual-Values} \\ 
 & Actual-Pos. & Actual-Neg. \\ \hline
Pos. Predicted. & TP & FP \\ \hline
Neg. predicted & FN & TN \\ \hline
\end{tabular}
\label{tab:3}
\end{table}

\subsubsection{Model Saving}\label{}

This stage is considered the last process in the batch processing phase, which includes saving the highest-performance model in the batch processing phase to use it as a predictive model in the real time streaming prediction phase.

\subsection{Real-Time Streaming Prediction Phase}
Our primary aim of the real-time Streaming prediction phase is to build a framework methodology to analyze the high velocity of streaming data arriving each second in real-time. Our methodology has four main components: data collection, data ingestion system, stream processing, and results visualization, as shown in Figure  \ref{fig1}. To check the proposed architecture's ability to identify suicidal ideation in real-time scenarios. We used Twitter API to retrieve real-time streaming tweets from Twitter. Twitter Streaming API \footnote{https://developer.twitter.com/en/docs/tutorials/consuming-streaming-data} is the basic method for accessing Twitter data. Twitter API allows access to real-time with a limited set of approximately 1\% of all tweets. Furthermore, Tweepy \footnote{https://docs.tweepy.org/en/latest/index.html} allows us to search tweets using hashtags, keywords, trends, geolocation, or timelines. Our methodology used keyword searches for retrieval of the tweets. We employed rules to retrieve only English tweets and filtered all duplicate tweets created by retweets. A total stream of 764 tweets was retrieved using multiple keywords related to suicidal ideation, including ``feel," ``want to die," and ``kill myself". The retrieved tweets included multiple columns, including tweet content, retweet counts, and usernames. Only the ``tweet" column was used for our work, while the other columns were not utilized and were removed from the collected data. Apache Kafka was utilized to develop real-time pipelines and stream data ingestion. The key benefit of Kafka is its ability to handle huge amounts of real-time data within low latency, and it is fault-tolerant and scalable to ingest large data streams. We created an input topic, ``Source-tweets," on the Kafka system.\\

The collected tweets were then ingested as data streams into the Apache Kafka input topic. Spark Structured Streaming consumes stream tweets from the Kafka topic in real-time into the unbounded table. We implemented several preprocessing steps to refine the tweets' stream effectively. These steps involve removing irrelevant information, reducing the noise, and extracting appropriate stream data. After preprocessing and cleaning the streaming tweets, we generated a feature vector and fed it into the highest accurate model previously developed and trained in the batch processing phase to predict suicidal ideation in real time. The prediction results were then pushed and buffered in a Kafka output ``Predicted-tweets" topic before being consumed by the Power BI application to visualize the final prediction results in real time.

\section{Experimental Setup and Performance Analysis}\label{sec4}

\subsection{Experimental Setup}\label{}

The proposed ApacheSpark-based architecture was implemented using the ``PySpark'' library to build the classification algorithms: NB, LR, LinearSVC, DT, RF, and MLP algorithms. Apache Spark Cluster was installed on a laptop with 64 GB of RAM, a 1 TB SSD disk drive, and an Intel Core i7 CPU (14 cores, 20 logical processors). In addition, we integrated multiple API libraries for implementation. ML library of Apache Spark was used to develop classification algorithms. Apache Kafka version of ``2.0.2'' was deployed as an input system for ingesting data streams from Twitter. Tweepy version of ``4.10.0'' for connecting to the Twitter API. Spark Structured Streaming was applied for receiving and processing stream tweets from Kafka topics—Power BI application for Visualizing the real-time streaming prediction results.

\subsection{Exploratory Data Analysis}\label{}

\begin{figure}[h!]
\centering
\begin{minipage}{0.49\textwidth}
\centering
\includegraphics[width=1\textwidth]{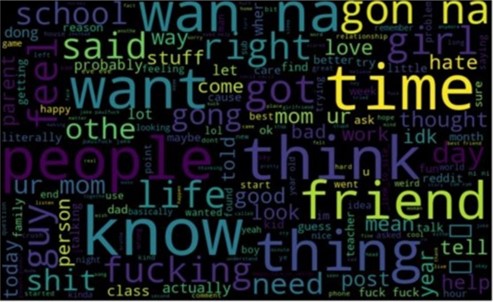}
\caption{Word cloud representation of suicidal-related postings}
\label{fig3}
\end{minipage}
\hfill
\begin{minipage}{0.49\textwidth}
\centering
\includegraphics[width=1\textwidth]{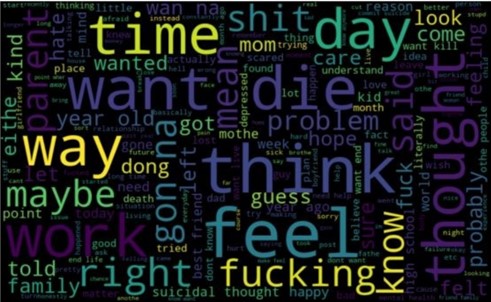}
\caption{Cord cloud representation of non-suicidal-related postings}
\label{fig4}
\end{minipage}
\end{figure}

We performed Word Cloud to explore the dataset utilized in this work. Word clouds serve as visual representations of the most frequently appeared terms within the dataset. The font size represents the frequency of each word within the dataset. From Figures \ref{fig3} and \ref{fig4}, We show that most suicidal postings contain the words ``want," ``friend," and ``think" in large letter or font sizes. The Word Cloud of nonsuicidal posts, on the other hand, larger fonts represent the most frequently nonsuicidal posts of repeated words, including the words ``though," ``feel" and ``die.''

\subsection{Evaluation of Batch Processing Phase}\label{}
\begin{figure}[h!]
\centering
\begin{minipage}{0.49\textwidth}
\centering
\includegraphics[width=1\textwidth]{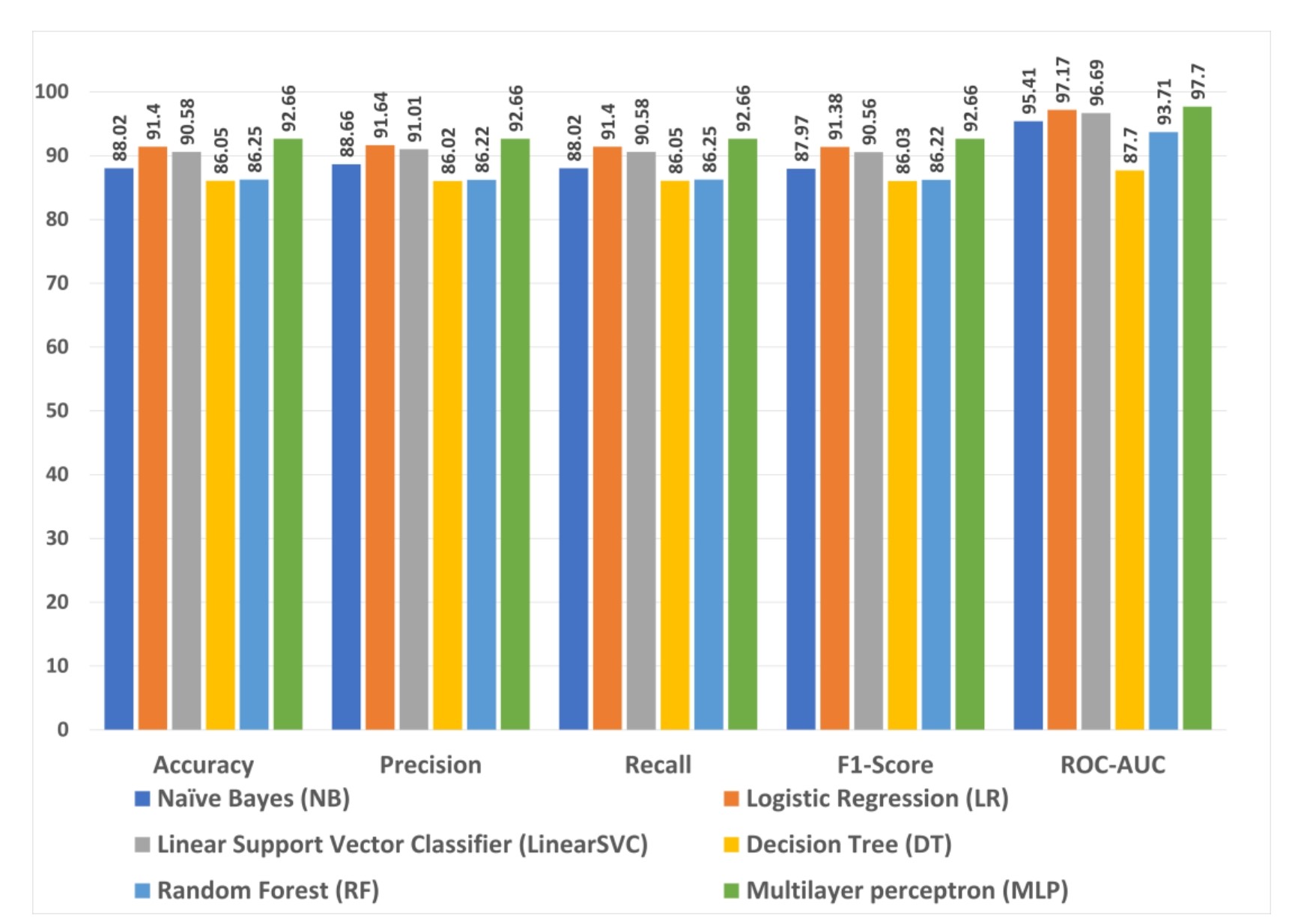}
\caption{Comparison of performance results of all classification algorithms with Unigram +TF-IDF features}
\label{fig5}
\end{minipage}
\hfill
\begin{minipage}{0.49\textwidth}
\centering
\includegraphics[width=1\textwidth]{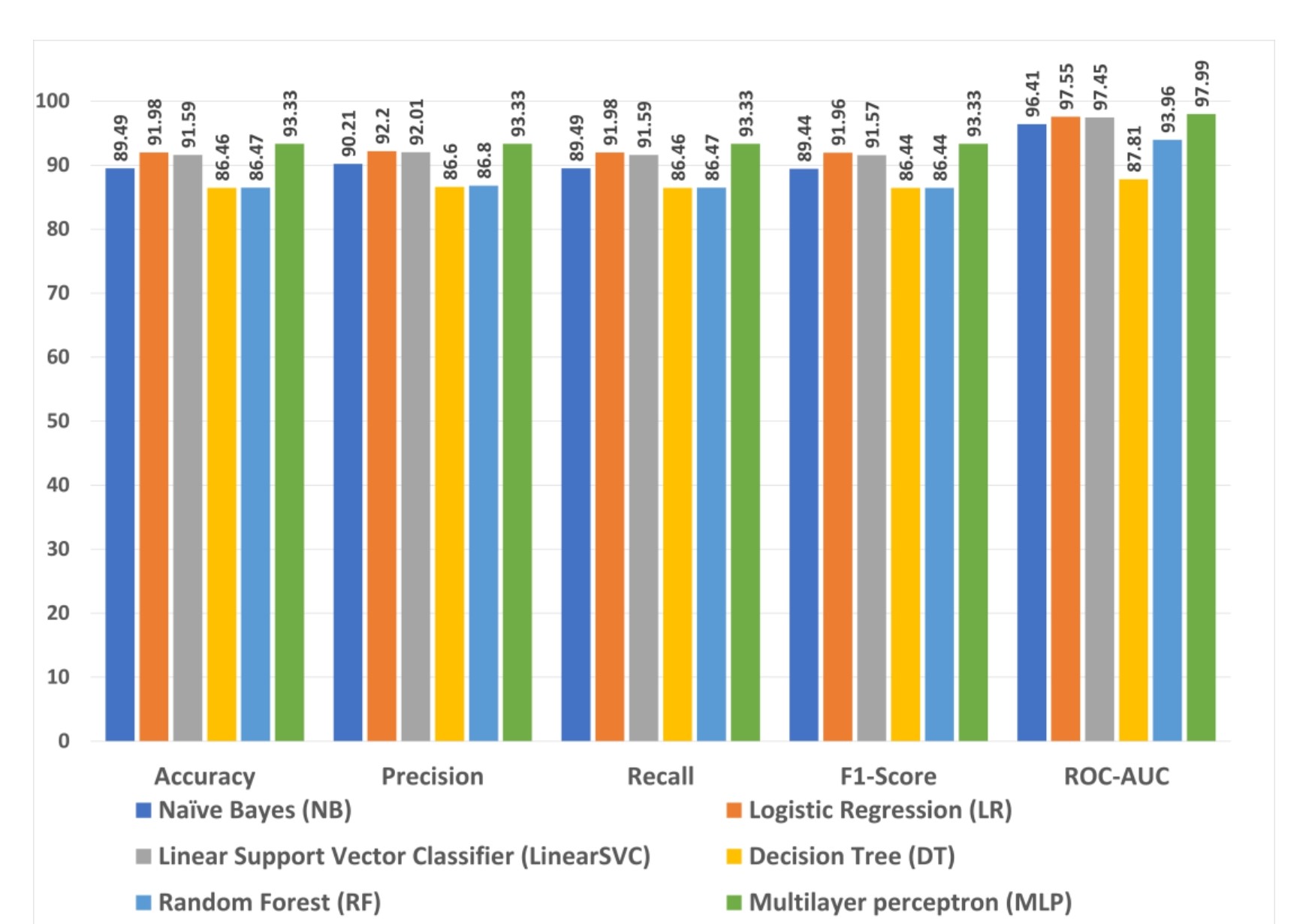}
\caption{Comparison of performance results of all classification algorithms with Unigram + CV-IDF features}
\label{fig6}
\end{minipage}
\end{figure}

\begin{figure}[h!]
\centering
\begin{minipage}{0.49\textwidth}
\centering
\includegraphics[width=1\textwidth]{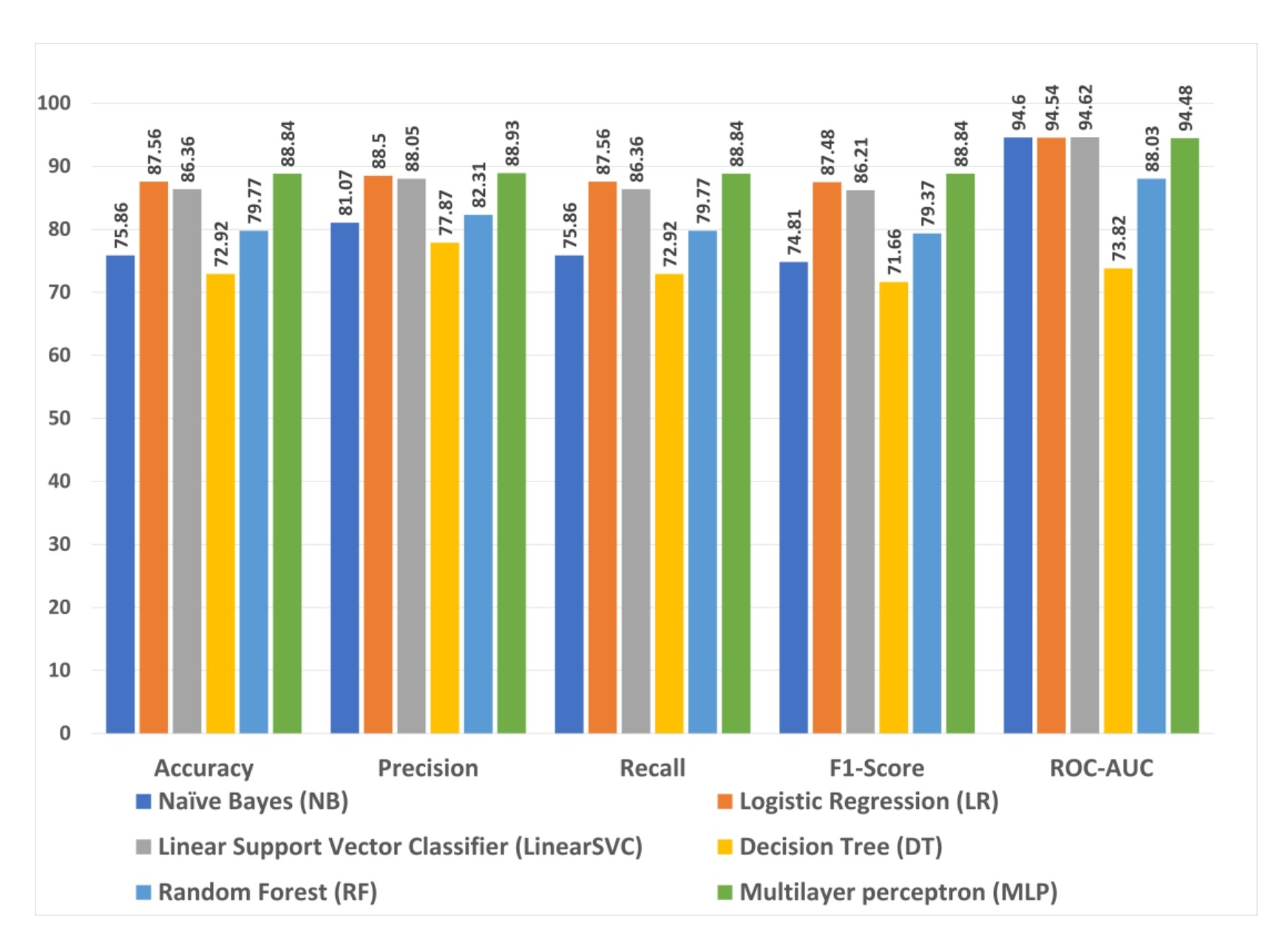}
\caption{Comparison of performance results of all classification algorithms with Bigram + CV-IDF features}
\label{fig7}
\end{minipage}
\hfill
\begin{minipage}{0.49\textwidth}
\centering
\includegraphics[width=1\textwidth]{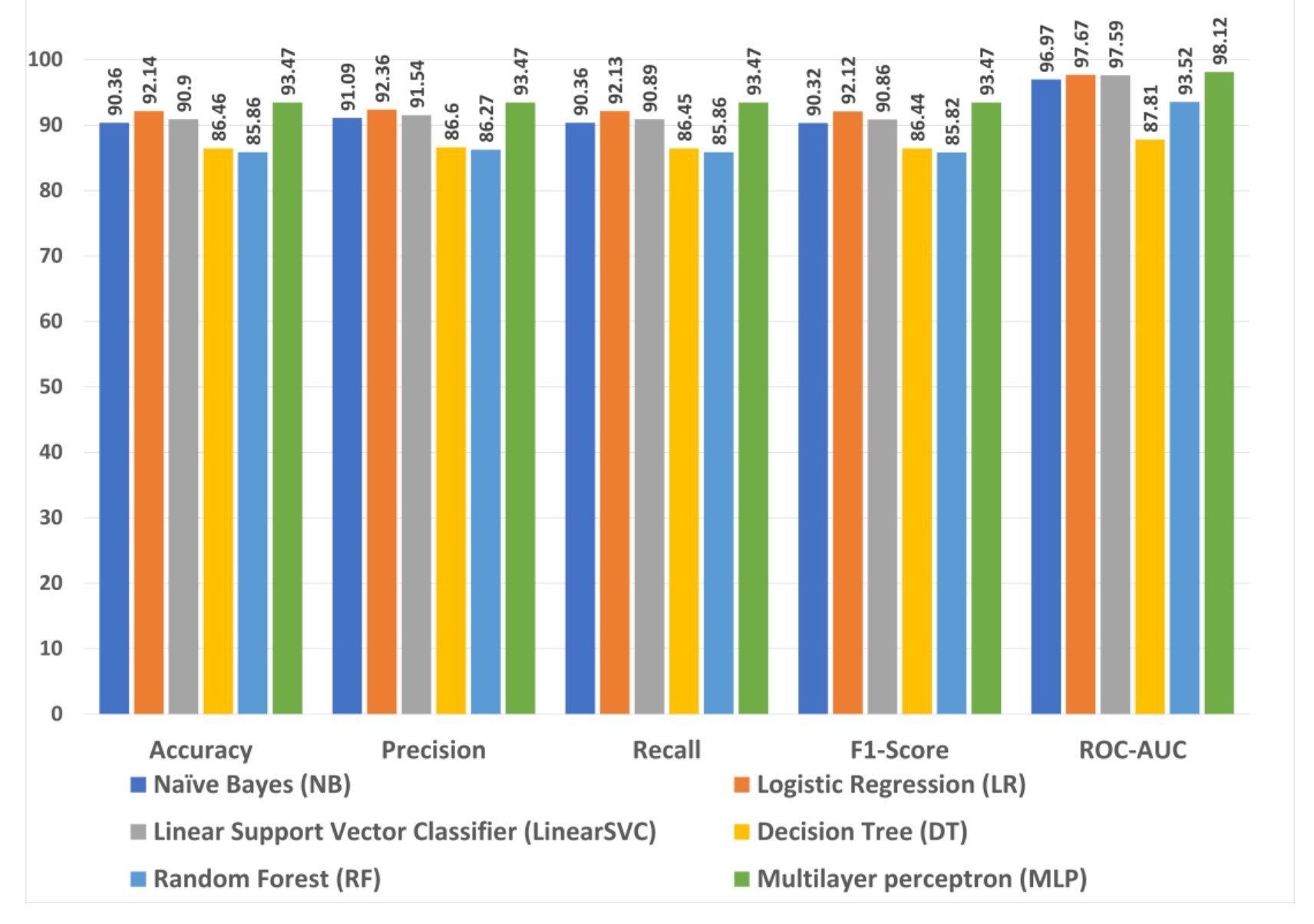}
\caption{Comparison of performance results of all classification algorithms with (Unigram + Bigram) + CV-IDF features}
\label{fig8}
\end{minipage}
\end{figure}

\begin{figure}[h!]
\centering
\begin{minipage}{0.49\textwidth}
\centering
\includegraphics[width=1\textwidth]{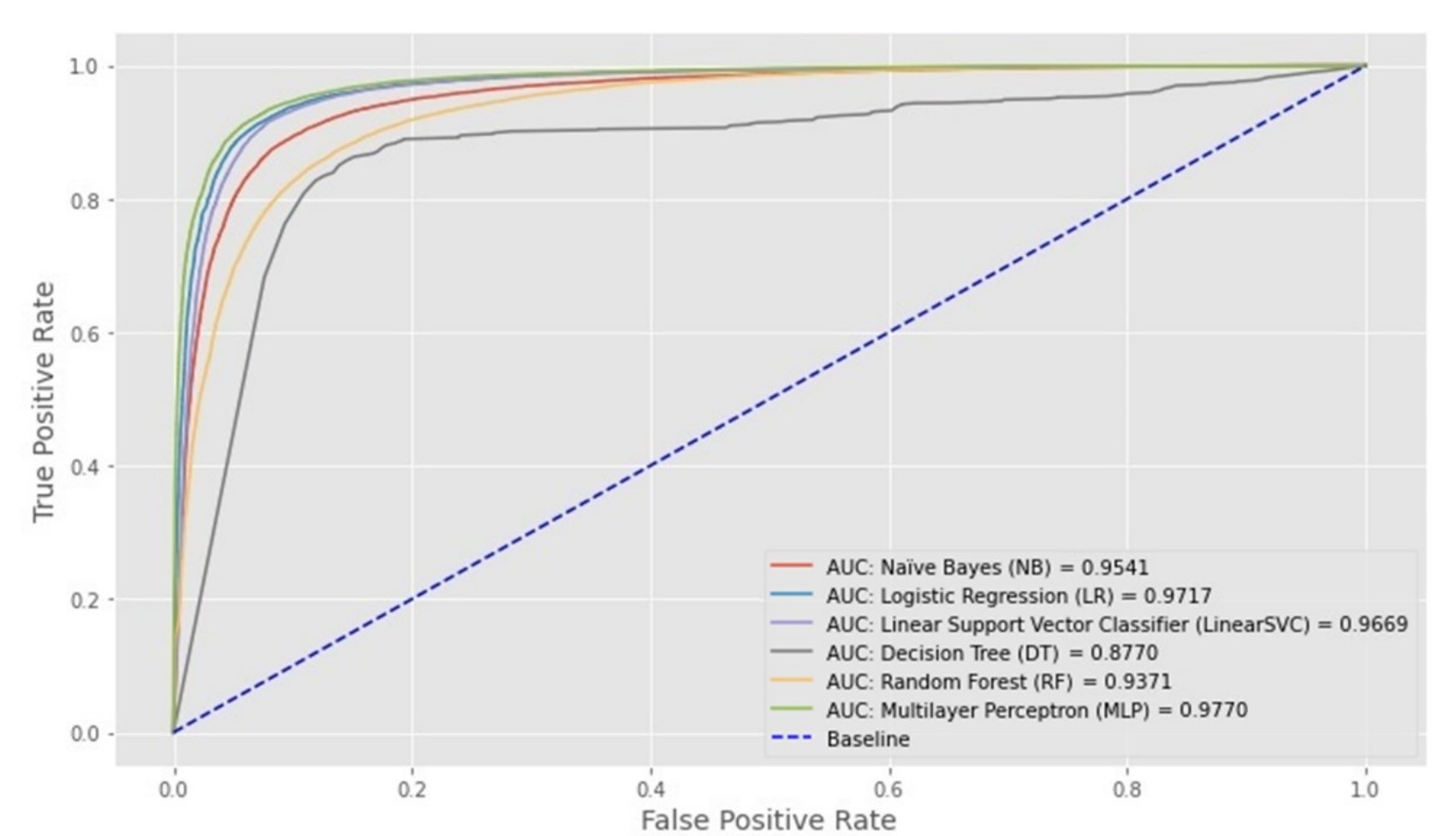}
\caption{Comparison of ROC-AUC of all classification algorithms with Unigram + TF-IDF features method}
\label{fig9}
\end{minipage}
\hfill
\begin{minipage}{0.49\textwidth}
\centering
\includegraphics[width=1\textwidth]{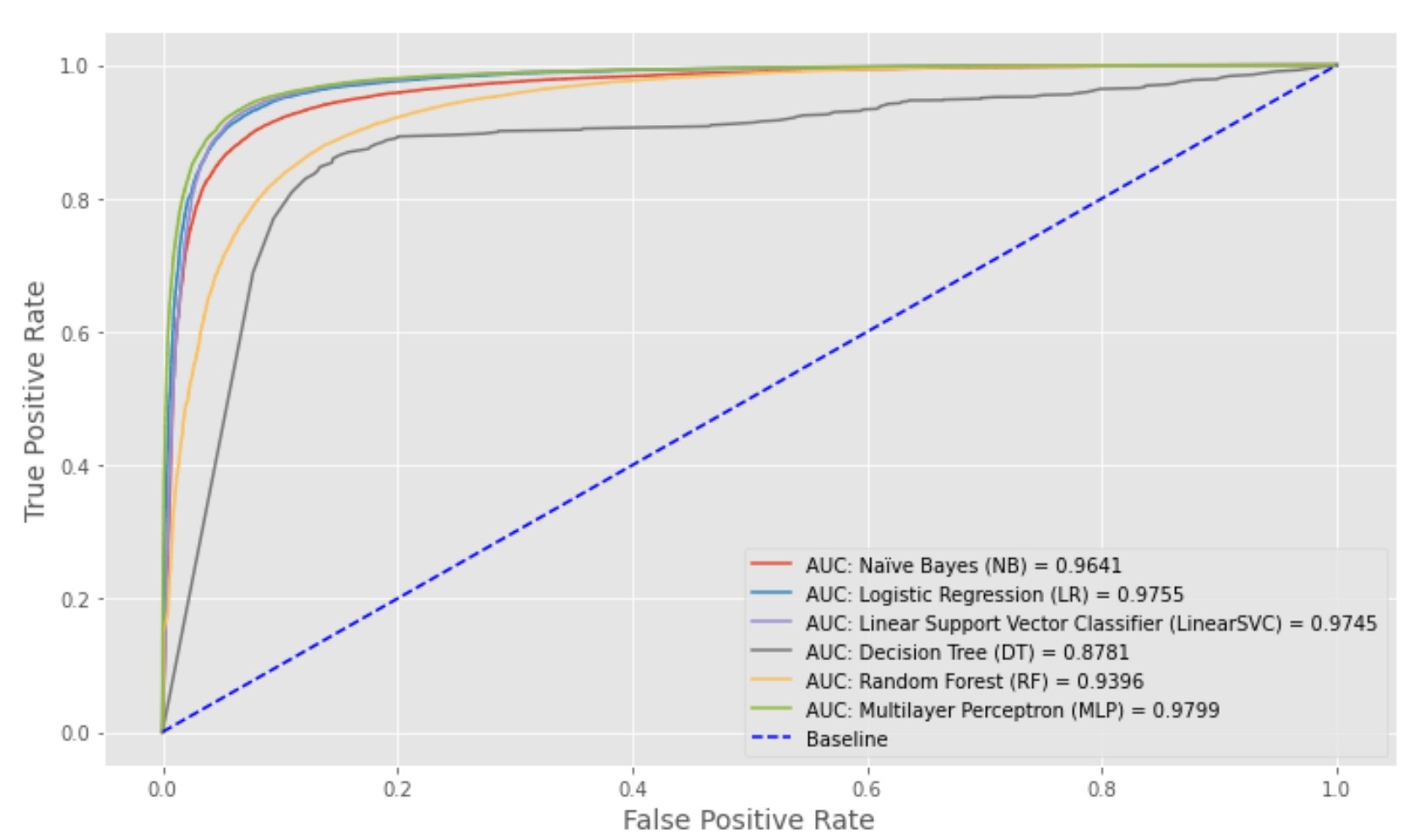}
\caption{Comparison of ROC-AUC of all classification algorithms with Unigram + CV-IDF features method}
\label{fig10}
\end{minipage}
\end{figure}

\begin{figure}[h!]
\centering
\begin{minipage}{0.49\textwidth}
\centering
\includegraphics[width=1\textwidth]{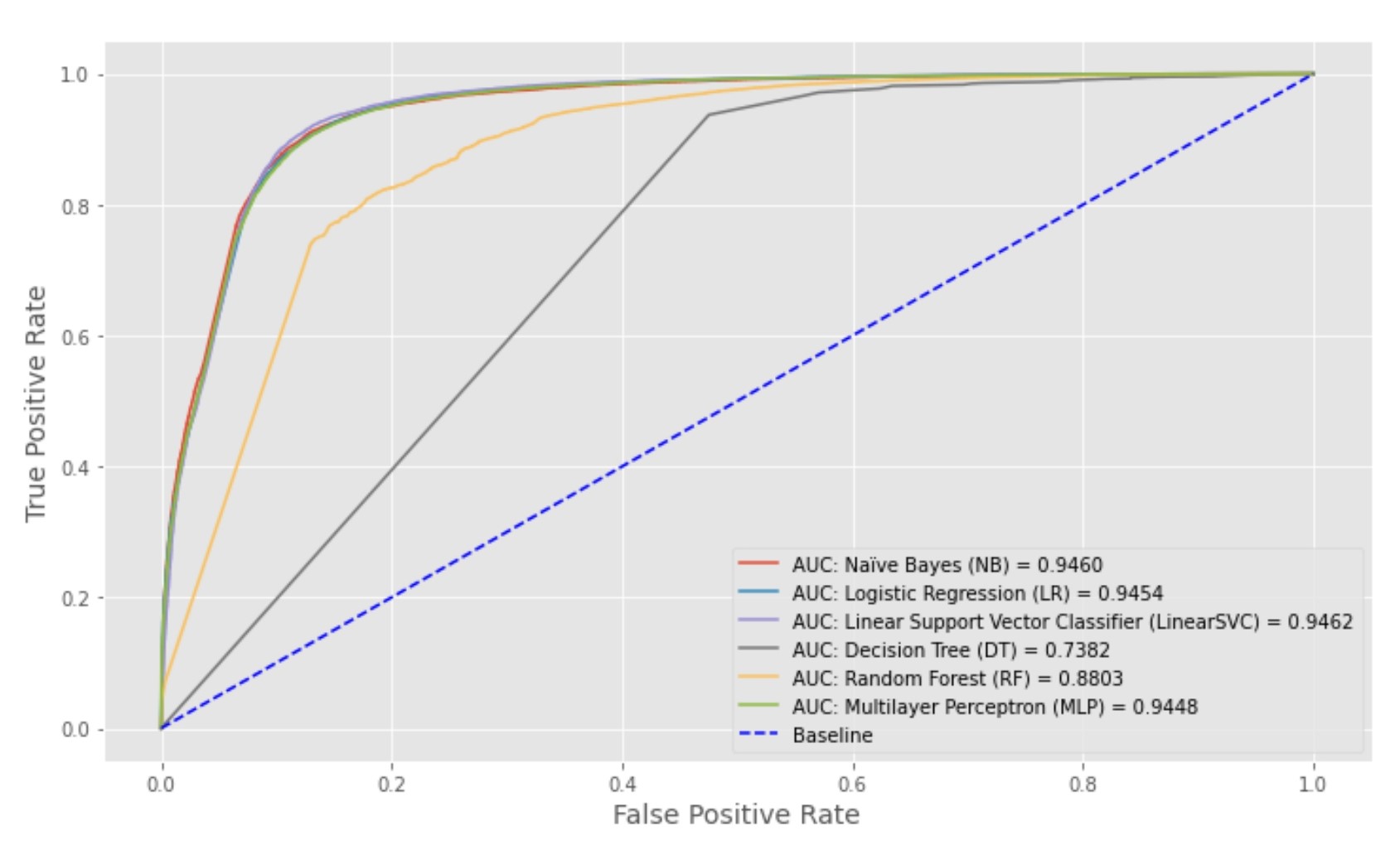}
\caption{Comparison of ROC-AUC of all classification algorithms with Bigram + CV-IDF features method}
\label{fig11}
\end{minipage}
\hfill
\begin{minipage}{0.49\textwidth}
\centering
\includegraphics[width=1\textwidth]{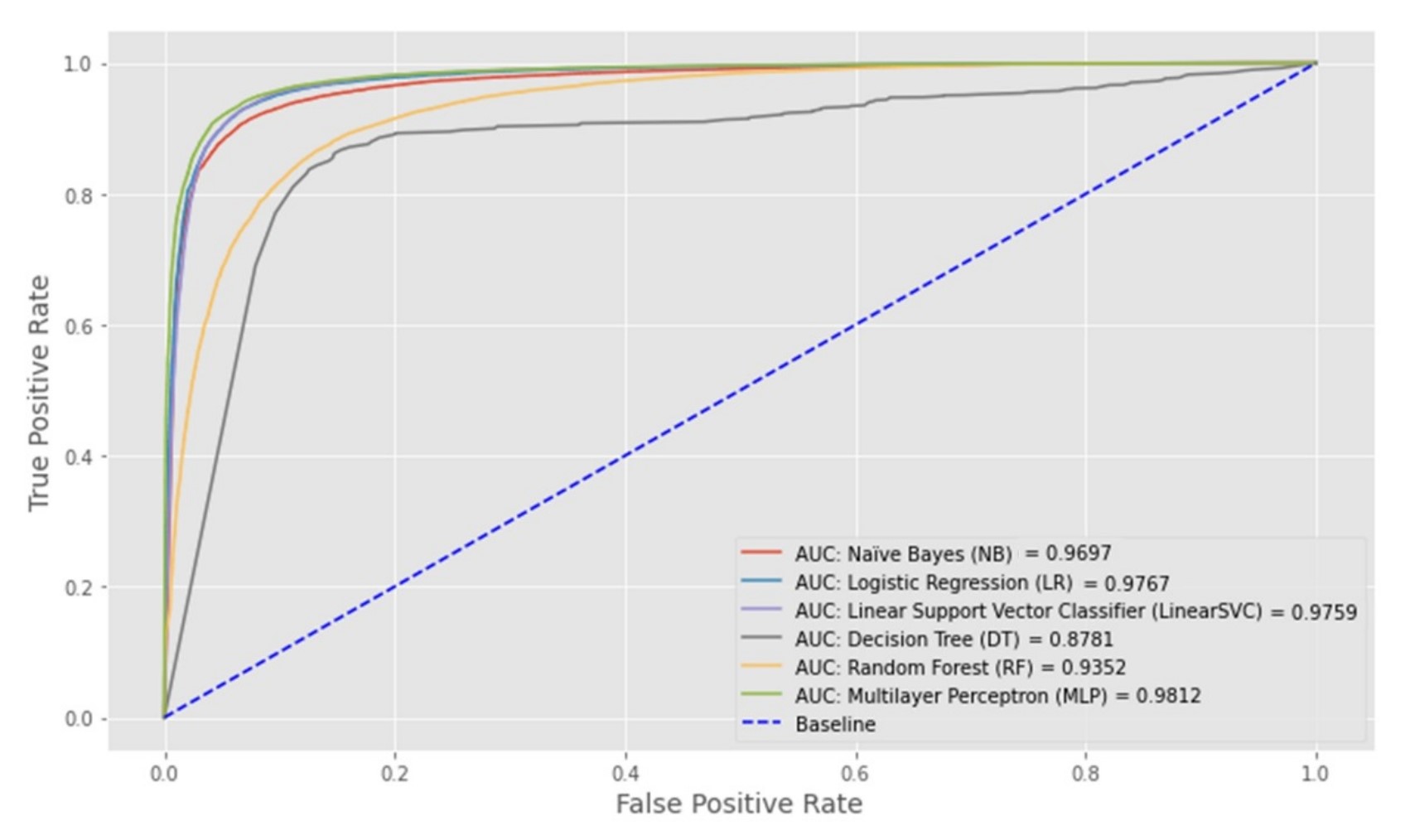}
\caption{Comparison of ROC-AUC of all classification algorithms with (Unigram + Bigram) + CV-IDF features method}
\label{fig12}
\end{minipage}
\end{figure}

This section presents and discusses the experimental applied in the batch processing phase for identifying suicidal ideation in individuals based on their social media posts. Our primary objective was to determine the most efficient model with the highest performance to adopt for real-time streaming prediction phase. We used multiple Apache Spark ML algorithms in this work, including Naïve Bayes (NB), Logistic Regression (LR), Linear Support Vector classifier (LinearSVC), Decision Tree (DT), Random Forest (RF), and Multilayer Perceptron (MLP). The algorithms were trained and evaluated using data from the Reddit forum, using three different strategies for feature extraction: TF-IDF, N-gram, and the CountVectorizer technique. Multiple combinations of these feature extraction methods were implemented to extract the essential features. A hyperparameter tuning strategy was adopted to detect the optimal parameter tune for each model configuration. Two methods are commonly employed for Hyperparameter tuning: Random search and Grid search.\\
In this work, we utilized the Grid search as a hyperparameter technique in the experiments. The Grid search hyperparameter tuning process aims to find the optimal parameters and most suitable values for each classifier to enhance the overall performance.\\
Furthermore, we made use of 10-fold Cross-validation, which is a widespread technique and reliable method for minimizing overfitting, enhancing the validity and reliability of the classification models, and balancing the bias and variance values. With the 10-fold cross-validation strategy, the given data were subdivided randomly into ten subsets of the same size; one subset was used for testing purposes, while the other nine subsets were used for the training process. Cross-validation was executed ten times, with each of the ten subsets used as validation only once. To get a final estimate, the data were averaged across ten folds. Table \ref{tab:4} and Figures \ref{fig5}, \ref{fig6}, \ref{fig7}, and \ref{fig8}   illustrate the experimental results and comparative performance assessment of multiple Spark ML classifiers using a binary classification evaluator.\\ 
From all experimental results, we found that the Multilayer Perceptron (MLP) classifier outperformed the other classification algorithms and achieved a greater accuracy rate of 93.47\% and an AUC socre of 98.12\%. The logistic Regression (LR) classifier also performed well but somewhat less than the Multilayer Perceptron (MLP) classifier and achieved the second-greatest performance, with an accuracy rate of 92.14\%. In addition, the results showed no significant performance difference between the Linear Support Vector classifier (LinearSVC) and Naïve Bayes (NB). Unexpectedly, from the experimental results, we found that Decision Tree (DT) and Random Forest (RF) underperformed other classifiers utilized in this work despite their efficacy in numerous machine-learning scenarios.\\ 
Also, from all experimental results, we have shown that most classifier models that used N-gram + CV-IDF as their feature extraction approach performed better than those that used the N-gram +TF-IDF feature approach. The classifier algorithms were also evaluated using another metric known as the Area-Under-Curve (AUC). The metric provides a value ranging from 0 to 1. A value closer to 1 indicated better classification results. Figures \ref{fig9}, \ref{fig10}, \ref{fig11} and \ref{fig12} display the AUC comparison of all the classification methods.

\begin{table}[h!]
\caption{Performance Comparison of Classification Algorithms on testing dataset}
\begin{tabular}{ccccccccc}
\hline

\textbf{Model} & \textbf{\textit{Feature Extraction Combination}}& \textbf{\textit{ACC.}}& \textbf{\textit{PRE.}} & \textbf{\textit{REC.}} & \textbf{\textit{F1.}} & \textbf{\textit{AUC.}}\\
\hline
    
\multirow{4}{*}{\textbf{NB}} 

& Unigram+TF-IDF  &88.02	&88.66	&88.02	&87.97	&95.41\\ 
& Unigram+CV-IDF  &89.49	&90.21	&89.49	&89.44	&96.41\\ 
& Bigram+CV-IDF  &75.86	&81.07	&75.86	&74.81	&94.60\\ 
& (Unigram + Bigram) + CV-IDF  &90.36	&91.09	&90.36	&90.32	&96.97\\ 
\hline

\multirow{4}{*}{\textbf{LR}} 
& Unigram+TF-IDF  &91.40	&91.64	&91.40	&91.38	&97.17\\ 
& Unigram+CV-IDF  &91.98	&92.20	&91.98	&91.96	&97.55\\ 
& Bigram+CV-IDF  &87.56	&88.50	&87.56	&87.48	&94.54\\ 
& (Unigram + Bigram) + CV-IDF  &92.14	&92.36	&92.13	&92.12	&97.67\\ 
\hline

\multirow{4}{*}{\textbf{LinearSVC}} 
& Unigram+TF-IDF  &90.58	&91.01	&90.58	&90.56	&96.69\\ 
& Unigram+CV-IDF  &91.59	&92.01	&91.59	&91.57	&97.45\\ 
& Bigram+CV-IDF  &86.36	&88.05	&86.36	&86.21	&94.62\\ 
& (Unigram + Bigram) + CV-IDF  &90.90	&91.54	&90.89	&90.86	&97.59\\ 
\hline

\multirow{4}{*}{\textbf{DT}} 
& Unigram+TF-IDF  &86.05	&86.02	&86.05	&86.03	&87.70\\ 
& Unigram+CV-IDF  &86.46	&86.60	&86.46	&86.44	&87.81\\ 
& Bigram+CV-IDF  &72.92	&77.87	&72.92	&71.66	&73.82\\ 
& (Unigram + Bigram) + CV-IDF  &86.46	&86.60	&86.45	&86.44	&87.81\\ 
\hline

\multirow{4}{*}{\textbf{RF}} 
& Unigram+TF-IDF  &86.25	&86.22	&86.25	&86.22	&93.71\\ 
& Unigram+CV-IDF  &86.47	&86.80	&86.47	&86.44	&93.96\\ 
& Bigram+CV-IDF  &79.77	&82.31	&79.77	&79.37	&88.03\\ 
& (Unigram + Bigram) + CV-IDF  &85.86	&86.27	&85.86	&85.82	&93.52\\ 
\hline

\multirow{4}{*}{\textbf{MLP}} 
& Unigram+TF-IDF  &92.66	&92.66	&92.66	&92.66	&97.70\\ 
& Unigram+CV-IDF  &93.33	&93.33	&93.33	&93.33	&97.99\\ 
& Bigram+CV-IDF & 88.84	&88.93	&88.84	&88.84	&94.48\\ 
& (Unigram + Bigram) + CV-IDF  &93.47	&93.47	&93.47	&93.47	&98.12\\ 
\hline
\end{tabular}
\label{tab:4}
\end{table}

\subsection{Evaluation of Real-Time Streaming Prediction Phase}\label{}

\begin{figure}[H]
\centering
\includegraphics[width=0.6\textwidth]{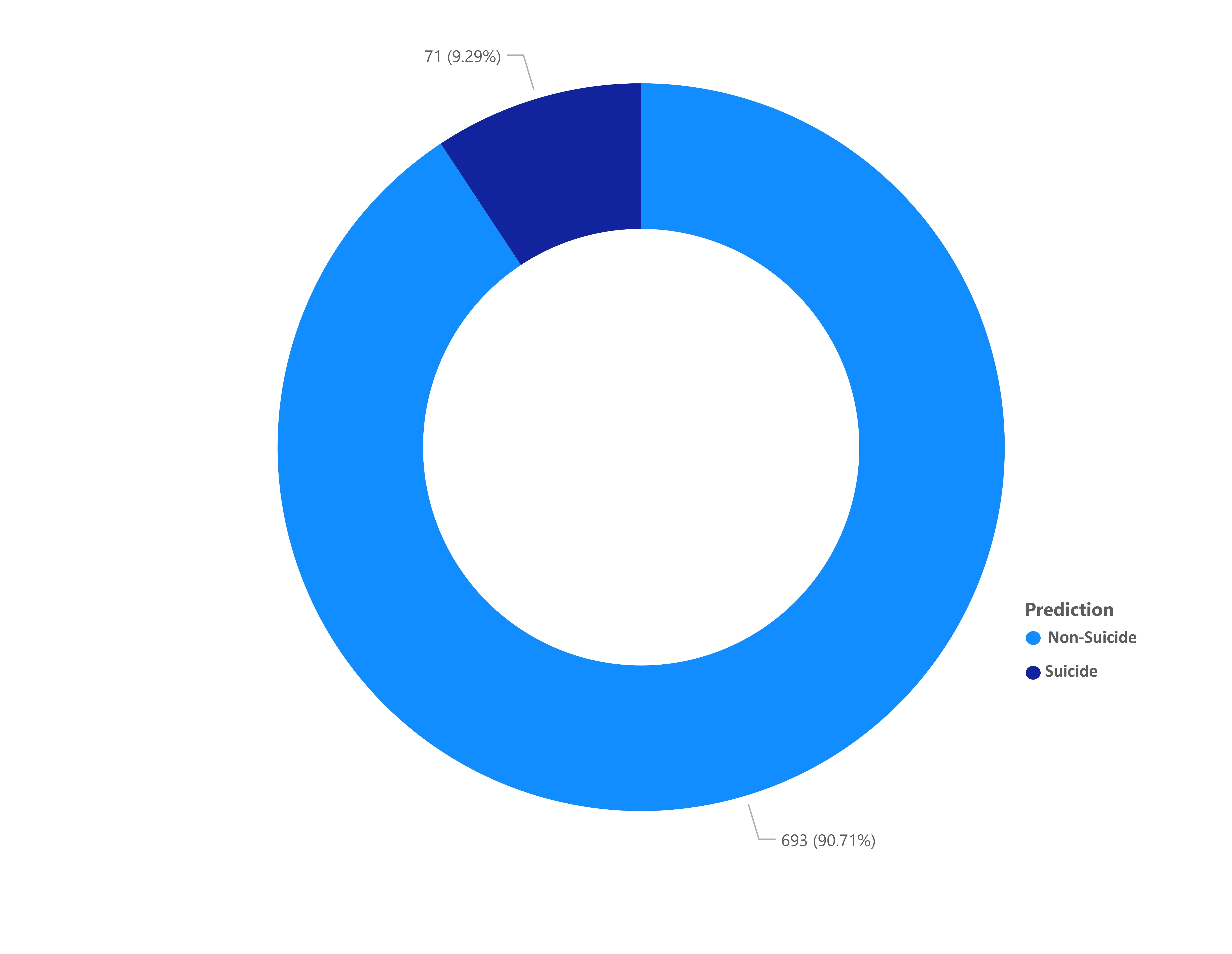}
\caption{Power BI visualization of real-time streaming prediction phase results}\label{fig14}
\end{figure}

The real-time streaming prediction phase used the classifier models already developed and pre-trained during the batch processing phase to evaluate their ability to predict suicidal ideation from Twitter streaming data. After designing and assessing the classifier models in the batch processing phase, the classifier with the greatest performance, as in our experiment, MLP with (Unigram + Bigram) + CV-IDF feature extraction combination, was applied for predicting Twitter suicidal ideation-related content in real-time. We collect streaming tweets using Twitter API with multiple keywords, including ``feel," ``want to die," and ``kill myself", which were then pushed into the Apache Kafka input topic. These streams of tweets are consumed by Apache Spark Structure Streaming from the Kafka input topic, which is then preprocessed as a data stream and used to generate a feature vector. The best pre-trained model already developed in the batch processing phase then analyzes the stream of preprocessed tweets and predicts whether these tweets are suicidal or normal content in real-time. The prediction results are then pushed to a Kafka output topic for buffering and then consumed from the Power BI application to visualize the prediction results in real-time. In our work, a total of 764 tweets as a data stream were collected to examine the prediction ability in the real-time streaming prediction phase. The real-time streaming prediction phase results indicated that (9.29\%) of the tweets were predicted as suicide, whereas (90.71\%) were non-suicide. Figure \ref{fig14} shows the results obtained in the real-time streaming prediction phase.

\section{Discussions}\label{sec5}

In this study, we proposed a big data approach to predict suicidal ideation based on data collected from social media platforms. The proposed methodology comprised two phases on batch processing and streaming predictions in real-time. The systems utilized six Spark ML algorithms to build the classification model and compared the performances of the models. In the streaming data pipeline, live streams of a tweet are collected from Twitter using the keywords ``feel", ``want to die" and ``kill myself" and then sent the collected data to the Kafka topic. Spark Structured Streaming receives the stream data from the Kafka topic, extracts the optimal feature, and then sends batches of preprocessed data to the real-time streaming prediction model to predict whether the tweet contains indications of suicidal ideation.

This work used three feature extraction methods, including TF-IDF, N-gram, and Count Vectorizer, with different combination scenarios to extract the optimal features from the input data. The experimental results of six classification models showed that the MLP classifier had the highest accuracy value of 93.47\% with the features extracted using (Unigram + Bigram) +CV-IDF feature extraction scenario. At the same time, a high accuracy of 93.33\% was obtained from the MLP classifier with features extracted using (Unigram + CV-IDF). In addition, MLP provided the best accuracy of 92.66\% using (Unigram + TF-IDF). 

In comparing our experimental results with related works, we noticed that the highest accuracy obtained from the MLP classifier is higher than XGBoost and logistic regression accuracies rate of 83.87\% and 86.45\%, respectively, achieved by S. Jain et al. \cite{rf9}. Also, compared with the accuracy and F1 score rate of 80\% and 92\%, respectively, achieved by A. E. Aladağ et al. \cite{rf13}. Furthermore, our methodology outperformed the accuracy rate of 76.80\% that was recorded by V. Desu et al. \cite{rf15}. In addition, our experimental results registered a higher performance than the Naïve Bayes algorithm, achieving a Precision value of 87.50\%, a Recall value of 78.8\%, and F1. value of 82.9\% by M. Birjali et al. \cite{rf21}. Therefore, we adopted the MLP classifier with (Unigram + Bigram) + CV-IDF feature combination scenario to predict suicidal ideation in the second phase of real-time streaming prediction using Twitter streaming data. 

That being said, further improvements can be made to extend this study. The first improvement can be achieved by increasing the number of features of the textual data using additional data such as emoticons, special characters, and symbols to extract optimal features and reduce the misclassification results. Moreover, the dataset can be expanded by gathering additional textual data from other social media platforms to make our data more representative and varied.

\section{Conclusion and Future work}\label{sec6}

In conclusion, this paper proposed a real-time streaming prediction system for suicidal ideation prediction of users' posts on social networks using a big data analytics environment—the work methodology analysis of social media content with two-phase batch processing and real time streaming prediction. Our system applied two types of datasets. Reddit's historical big data are used for model building, while Twitter streams big data have been used for real-time streaming prediction. 
Our proposed methodology for building binary classification models was evaluated using various assessment metrics and showed high levels of accuracy and AUC scores with stable Recall and Precision. 
The experimental results of the batch processing phase revealed that the MLP classifier achieved the highest classification accuracy of 93.47\% on an unseen dataset and was used for the real-time streaming prediction phase. According to the results of various testing scenarios, we can conclude that the features retrieved from stream data could accurately determine the suicidal ideation of users in real time. The developed system might also assist public health professionals with limited resources in determining and controlling suicidal ideation and preparing preventative steps to save lives. Multiple languages, such as Turkish and Arabic, can be added for future work. To deal with such datasets, which require sequential information and local feature engineering, we may use Ensemble LSTM and CNN models for better performance. We also plan to develop a web or mobile interface as a text-analysis tool to detect the individual's health status.

\bibliographystyle{sn-aps}     
\bibliography{sn-article} 
\end{document}